\documentclass[lettersize,journal]{IEEEtran}
\usepackage{amsmath,amsfonts}
\usepackage{algorithmic}
\usepackage{algorithm}
\usepackage{array}
\usepackage[caption=false,font=normalsize,labelfont=sf,textfont=sf]{subfig}
\usepackage{textcomp}
\usepackage{stfloats}
\usepackage{url}
\usepackage{verbatim}
\usepackage{graphicx}
\usepackage{cite}
\usepackage{cleveref}
\usepackage{booktabs}
\usepackage{multirow}
\usepackage{authblk}

\begin{document}

\title{TIRAuxCloud: A Thermal Infrared Dataset for Day and Night Cloud Detection}

\author[1,3]{Alexis Apostolakis}
\author[1,2]{Vasileios Botsos}
\author[4]{Niklas Wölki}
\author[4]{Andrea Spichtinger}
\author[2]{Nikolaos Ioannis Bountos}
\author[2,3]{Ioannis Papoutsis}
\author[1]{Panayiotis Tsanakas}


\affil[1]{National Technical University of Athens, School of Electrical and Computer Engineering}
\affil[2]{School of Rural, Surveying and Geoinformatics Engineering, National Technical University of Athens}
\affil[3]{National Observatory of Athens}
\affil[4]{OroraTech GmbH}




\maketitle

\begin{abstract}
Clouds are a major obstacle in Earth observation, limiting the usability and reliability of critical remote sensing applications such as fire disaster response, urban heat island monitoring, and snow and ice cover mapping. Therefore, the ability to detect clouds 24/7 is of paramount importance. While visible and near-infrared bands are effective for daytime cloud detection, their dependence on solar illumination makes them unsuitable for nighttime monitoring. In contrast, thermal infrared (TIR) imagery plays a crucial role in detecting clouds at night, when sunlight is absent. Due to their generally lower temperatures, clouds emit distinct thermal signatures that are detectable in TIR bands. Despite this, accurate nighttime cloud detection remains challenging due to limited spectral information and the typically lower spatial resolution of TIR imagery. 
To address these challenges, we present \textit{TIRAuxCloud}, a multi-modal dataset centered around thermal spectral data to facilitate cloud segmentation under both daytime and nighttime conditions.  The dataset comprises a unique combination of multispectral data (TIR, optical, and near-infrared bands) from Landsat and VIIRS, aligned with auxiliary information layers. Elevation, land cover, meteorological variables, and cloud-free reference images are included to help reduce surface-cloud ambiguity and cloud formation uncertainty. To overcome the scarcity of manual cloud labels, we include a large set of samples with automated cloud masks and a smaller manually annotated subset to further evaluate and improve models. 
Comprehensive benchmarks are presented to establish performance baselines through supervised and transfer learning, demonstrating the dataset’s value in advancing the development of innovative methods for day and night time cloud detection.
\end{abstract}


\section{Introduction}

\label{sec:intro}
Accurate and consistent cloud detection remains a fundamental requirement for a wide range of remote sensing applications not only during the day but during the night as well. The presence of clouds acts as a major obstacle to satellite-based observation, as they obscure the retrieval of surface and atmospheric information \cite{li_cloud_2022, VIIRSnight,from_ground_infrared_2025, nasacloud, review_cloud_2020}.
As a result, several critical remote sensing services could underperform or send misleading signals to end users, causing them to act (or not act) on faulty data without being aware of it. Notable examples of remote sensing services that are needed to operate day and night are \textit{fire detection, monitoring and assessment}, \textit{precipitation forecasting},\textit{ urban heat island}, \textit{surface temperature},\textit{ snow and ice cover mapping},  \textit{aerosol optical depth} \cite{brazil_clouds_2008, monitoragri,li_cloud_2022, hyperspectral}. While cloud detection methods have seen significant advancements, particularly for daytime remote sensing images, nighttime cloud detection presents substantial challenges \cite{VIIRSnight, nasacloud,night_cloud_2020,swincloud,nature_cloud_2024}. During the day, visible and near infrared (VNIR) bands are effective due to the high solar reflectivity of clouds. However, the absence of direct solar illumination at night makes these bands unsuitable \cite{night_cloud_2020,swincloud}. 
On the other hand, thermal infrared (TIR) radiation is continuously emitted and can be observed regardless of solar illumination using satellite TIR sensors from satellite missions such as Landsat, VIIRS, and MODIS \cite{barsi2014landsat,modisviirs}, or even with ground-based TIR sensors \cite{from_ground_infrared_2025}. Clouds emit thermal energy proportional to their temperature, and they are generally colder than the Earth's surface or ocean temperatures. This temperature difference allows for their detection using thermal imagery from satellites in wavelengths such as the 8.0-12.0 $\mu$m window region (LWIR) \cite{nasacloud}. A significant challenge that arises when relying on thermal infrared data is distinguishing clouds from snow and ice, because both are cold and have similar temperatures \cite{li_cloud_2022}. Existing methods (either based on traditional thresholding \cite{scaramuzza, ThresAlgo, fmask_2023} or more advanced deep learning approaches \cite{cdnetv2, cloudu-net, swincloud, from_ground_infrared_2025, HRCloudnet}) struggle in night-time scenarios. 

\begin{figure*}[htbp]
    \centering
    \includegraphics[width=0.9\linewidth]{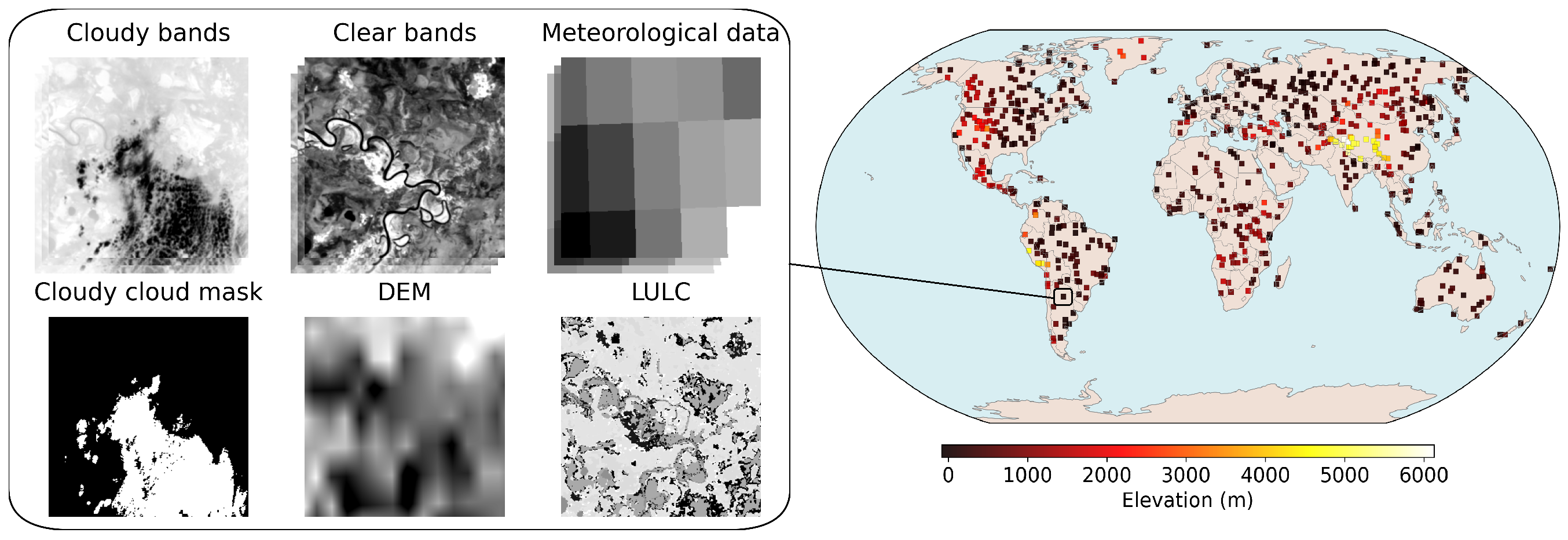}
    \caption{Global distribution of sampled Landsat scenes used for training, colored by elevation. Each one contains all the cloudy bands mentioned in \Cref{sec:dd}  alongside their clear counterparts and the auxiliary bands. Cloudy bands panel displays Cloudy Radiance B10/B11 mean, while the Clear bands panel shows Clear Radiance B10/B11 mean.}
    \label{fig:landsat_sampling}
\end{figure*}

A significant obstacle hindering accurate all-day cloud detection has been the absence of large, publicly available datasets specifically designed for cloud detection, using data available during day and night such as thermal emission observations \cite{nature_cloud_2024, swincloud, hyperspectral}. However, cloud properties are controlled by a range of environmental factors that are not fully captured in the thermal signal alone. Some studies have shown that incorporation of auxiliary information, such as Digital Elevation Model (DEM), Land Use Land Cover (LULC) or meteorological values, can provide valuable contextual cues for distinguishing clouds from cold surfaces such as snow, ice or high elevation terrain. In particular, DEM has shown to improve cloud and snow separation in mountainous regions \cite{wu2021geoinfoNet, li_cloud_2022} and atmospheric state variables affect cloud formation and thickness \cite{haynes2022low}. Despite these indications, current public datasets rarely include such auxiliary layers and are typically limited to spectral imagery alone, as seen in (Table~\ref{tab:cldatasets}). 

To address these gaps, we introduce \textit{TIRAuxCloud}, a novel dataset for cloud detection covering day and night conditions, primarily based on thermal infrared data. TIRAuxCloud combines imagery from established satellite missions, specifically Landsat 8/9 and VIIRS, known for their thermal infrared capabilities. We provide all spectral bands from both satellites enabling studies for day and night conditions. The dataset includes 21,058 Landsat image samples labeled by the automated Landsat algorithm \cite{fmask_2023}, 2,563 images extracted from manually annotated scenes from a USGS dataset \cite{scaramuzza_landsat8_col1_2021, scaramuzza_landsat8_col2_2021} and 5,004 VIIRS images created from scenes downloaded from World Bank \cite{worldbank_len_2021} collections. Our approach extends beyond raw satellite imagery by integrating auxiliary data to investigate the potential to enhance cloud detection accuracy, under conditions where reflectance spectral bands are unavailable.  These auxiliary data, chosen for their availability during both day and night, include i) \textit{Cloud-free counterparts} precisely co-registered with cloudy scenes, serving as clear-sky references; ii) \textit{DEM} useful for improving thin or broken cloud detection, especially in complex terrain \cite{nature_cloud_2024}; iii) \textit{Weather} parameters from Copernicus ERA5 \cite{hersbach2020era5} such as temperature, dew point temperature, surface pressure and precipitation that significantly influence cloud formation, cloud types and states \cite{nasawang2020,from_ground_infrared_2025}; iv) \textit{LULC maps} which help capture how different surfaces emit thermal radiation, aiding cloud detection over challenging backgrounds like snow, bare land, and artificial lights \cite{nature_cloud_2024,swincloud}; and finally v) \textit{Metadata} from Landsat and VIIRS such as Geolocation, Acquisition Date and Time, Sun Elevation and Azimuth Angles, which can serve as additional model inputs  or support statistical analyses \cite{nasacloud, viirsnn, night_cloud_2020, allclear}

By providing this rich, multi-modal dataset, we aim to overcome the limitations of current training data for all-day cloud segmentation, paving the way for more robust and accurate deep learning or foundation models capable of performing the task. Our contributions are 
i) the integration of, to the best of our knowledge, the most comprehensive set of auxiliary data available for day and night cloud detection offering rich contextual information to support learning;
ii) the benchmarking with widely adopted architectures and models tailored for cloud detection, establishing strong baselines for future research; 
iii) the provision of strong indications that auxiliary data can enhance model performance; and iv) evidence that the main automatically annotated Landsat dataset offers sufficiently rich information to support robust model training, even more than the smaller manually annotated Landsat dataset.

\subsection{Related Work}
\label{sec:rw}

Deep learning has enabled effective cloud segmentation, with many models using encoder-decoder architectures, some tailored to the specific challenges of the task \cite{review_cloud_2020,li_cloud_2022, cloudu-net, from_ground_infrared_2025, swincloud, benchcloudvision}. CDnetV2  \cite{cdnetv2} and Cloud-Adapter \cite{cloud-adapter} are two notable examples with open-source PyTorch implementations. CDnetV2 focuses on cloud-snow coexistence using a CNN encoder-decoder with adaptive feature fusion and semantic guidance. Cloud-Adapter leverages pre-trained Vision Foundation Models, enhancing segmentation with minimal extra parameters.

Recent work has also focused on the use of thermal infrared data for all-day cloud segmentation. \cite{nature_cloud_2024} introduces a unique approach to nighttime satellite cloud detection by generating machine learning training samples from daytime thermal infrared data, exploiting the consistent thermal characteristics of clouds across day and night. SwinCloud \cite{swincloud} introduces, for the first time, a Swin-UNet for cloud detection, integrating a Global–Local Swin Transformer block (GLST) and a Feature Fusion Module (FFM) for robust, multi-scale cloud detection in thermal infrared imagery. Other models, such as MFFCD-Net and CloudU-Net variants, address day/night detection challenges using VIIRS Day/Night Band and ground-based TIR camera images as input, respectively \cite{VIIRSnight, from_ground_infrared_2025}.

\begin{table*}[ht]
\footnotesize
    \centering
    \caption{Overview of publicly available remote sensing datasets for cloud detection}
    \label{tab:cldatasets}
    \begin{tabular}{@{}lllrrlll@{}}
        \toprule
        \textbf{\multirow{2}{*}{Dataset}} & \textbf{\multirow{2}{*}{Source}} & \textbf{Bands} & \textbf{Samples} & \textbf{\multirow{2}{*}{Size}} & \textbf{Resolution} & \textbf{\multirow{2}{*}{Mask}} & \textbf{\multirow{2}{*}{Notes}} \\
        & & \textbf{Auxiliary} & \textbf{Number} & & (Resampled) & & \\
        \midrule
        38/95-Cloud & Landsat 8/9 & VNIR & 34,701 & 384px & 30\,m & Manual & No TIR \\
        \midrule
        CloudSEN12+ & Sentinel 1/2 & VNIR, SAR, DEM, LULC & 50,249 & 509px, 2000px & 10\,m & Manual & No TIR \\
        \midrule
        \multirow{2}{*}{AllClear} & Sentinel 1/2, & VNIR, SAR, DEM, & \multirow{2}{*}{4,354,652} & \multirow{2}{*}{256px} & \multirow{2}{*}{10\,m} & \multirow{2}{*}{Auto} & Cloud removal focus; \\
                 & Landsat 8/9 & LULC, Metadata & & & & & small TIR patches \\
        \midrule
        \multirow{2}{*}{SPARCS} & \multirow{2}{*}{Landsat 8} & TIR, VNIR, & \multirow{2}{*}{80} & \multirow{2}{*}{1000px} & \multirow{2}{*}{30\,m} & \multirow{2}{*}{Manual} & TIR capable; no auxiliary; \\
               & & Metadata & & & & & limited samples \\
        \midrule
        \multirow{2}{*}{L8 Biome} & \multirow{2}{*}{Landsat 8} & \multirow{2}{*}{VNIR, TIR} & \multirow{2}{*}{192} & \multirow{2}{*}{100,000 m (scenes)} & \multirow{2}{*}{30\,m} & \multirow{2}{*}{Manual} & TIR capable; no auxiliary; \\
        & & & & & & & moderate samples \\
        \midrule
        \multirow{2}{*}{TIRAuxCloud} & Landsat 8/9, & TIR, VNIR, DEM, & \multirow{2}{*}{28,625} & \multirow{2}{*}{256px} & Landsat 100m & Auto \& & TIR with auxiliary data; \\
        & VIIRS & Met., LULC, Metadata & & & VIIRS 200m & Manual & sufficient samples \\
        \bottomrule
    \end{tabular}
\end{table*}

Several publicly available datasets (Table~\ref{tab:cldatasets}) support cloud detection and related tasks, mainly relying on reflectance bands (e.g., visible and near-infrared) from satellite imagery. The \textit{38-Cloud} \cite{38-cloud-1} and \textit{95-Cloud} \cite{95-cloud} datasets consist of 38 and 95 manually annotated Landsat 8/9 scenes, respectively, providing a total of 70,404 image patches. However, these datasets do not include processed thermal bands from Landsat imagery. The \textit{CloudSEN12} \cite{aybar2022cloudsen12} dataset and its extension, \textit{CloudSEN12+} \cite{aybar2024cloudsen12+}, are large, globally distributed collections based primarily on Sentinel-2 imagery, offering 49,400 image patches for the semantic segmentation of clouds and cloud shadows. The dataset includes high-quality manual annotations for thick clouds, thin clouds, and cloud shadows, along with Sentinel-2 spectral bands, co-registered Sentinel-1 SAR imagery, and auxiliary data such as DEMs and LULC maps. However, it lacks thermal infrared imagery, limiting its applicability to daytime conditions only. \textit{AllClear} \cite{allclear} is the largest public dataset for cloud removal, comprising 23,742 globally distributed regions and over 4 million images. It includes multispectral imagery from Sentinel-1/2 and Landsat-8/9, with thermal bands from Landsat. However, the resampling of all bands to 10m to match Sentinel-2 optical bands results in small patches (~2.5km), limiting suitability for cloud segmentation using Landsat thermal data, which has 100m native resolution. \textit{SPARCS} is a widely used public dataset for identifying clouds and cloud shadows, consisting of 80 Landsat-8 scenes, however it  is relatively small in size, which limits its suitability for training robust deep learning cloud detection models that rely on thermal information. The \textit{L8 Biome} dataset \cite{scaramuzza, scaramuzza_landsat8_col1_2021, scaramuzza_landsat8_col2_2021} is one of the few manually annotated resources that include thermal data (Landsat bands 10 and 11), though it is provided as full scenes requiring preprocessing for model training. It is also limited in size, probably reducing the model’s ability to generalize.

Landsat 8/9 and several other satellite missions contribute thermal infrared (TIR) data for cloud detection under both day and night conditions. Landsat’s automated cloud masks are generated using the pixel-based Fmask algorithm, for which up to 96\% accuracy has been reported \cite{fmask_2023}. VIIRS (Visible Infrared Imaging Radiometer Suite) offers 16 moderate-resolution channels used in neural network-based cloud masks (e.g., NNCM), along with VIIRS/CrIS fusion-derived features approximating MODIS-like channels. SDGSAT-1, via its Thermal Infrared Imager (TIS), provides three TIR bands (B1–B3) and is commonly used to evaluate model transferability across sensors. MODIS and Himawari-8 supply multi-spectral coverage, including key TIR wavelengths (8–14$\mu$m), enabling reliable cloud detection in nighttime imagery. Finally, AVHRR/HRPT supports day-night cloud detection using infrared channels 3–5, without reliance on visible light, making it effective under low-illumination conditions.

\section{Dataset Description}
\label{sec:dd}

To support the development of accurate cloud detection models capable of operating under varying illumination and surface conditions, we construct TIRAuxCloud\footnote{https://github.com/Orion-AI-Lab/TIRAuxCloud, https://huggingface.co/datasets/tirauxcloud/TIRAuxCloud}, a dataset compiled from multiple satellite sources, centered on thermal infrared (TIR) observations and enriched with auxiliary geospatial layers. The dataset is designed to address limitations in existing resources by providing structured, spatially aligned  data suitable for both daytime and nighttime segmentation tasks. The dataset is organized into three distinct subsets: the \textit{Landsat Main} Dataset, the Manually Annotated (MA) Landsat Dataset, and the VIIRS Dataset. By integrating thermal observations with auxiliary surface and atmospheric information, TIRAuxCloud supports robust cloud segmentation across a wide range of geographic regions, temporal variations, and illumination conditions. 

\subsection{Landsat Main Dataset}

The \textit{Landsat Main} subset is the primary component of TIRAuxCloud. It is designed to provide a reliable and representative foundation for supervised cloud segmentation. Leveraging thermal infrared channels in conjunction with reflective bands and geospatial context, this subset supports learning under diverse surface, atmospheric, and illumination conditions. A key objective in constructing this set was to provide Landsat sensor data, alongside high quality auxiliary layers. This enables TIR-based models to better distinguish clouds from cold surfaces such as snow/ice or cold mountain tops, particularly during nighttime. Similarly, it can help VNIR-based models distinguish bright surfaces (e.g. snow, urban areas) from clouds during the day. 
Each cloudy scene is paired with a clear counterpart, providing a cloud-free surface reference that helps distinguish ambiguous or semi-transparent clouds. 

\begin{figure}[htbp]
    \centering
    \includegraphics[width=1\linewidth]{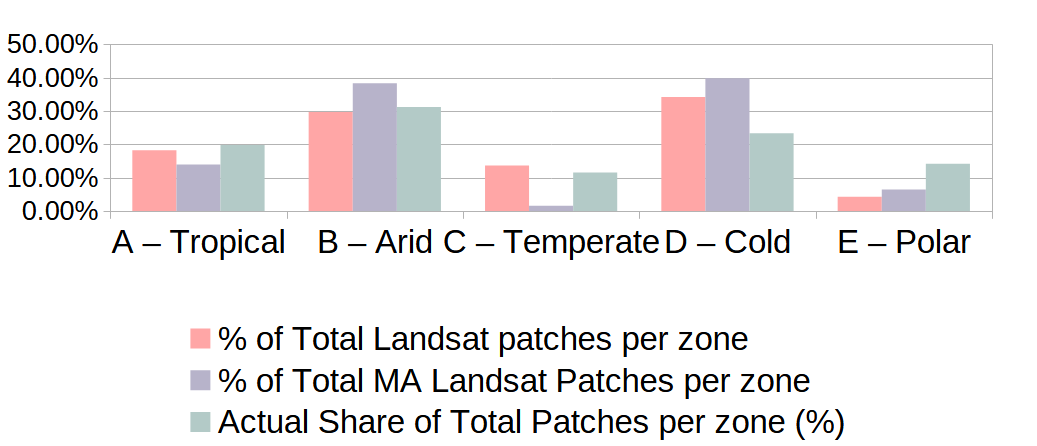}
    \caption{Distribution of Landsat and Manually Annotated Landsat patches across Köppen–Geiger climate zones \cite{beck2018koppen}.}
    \label{fig:climate_zones}
\end{figure}

All \textit{\textit{Landsat Main}} scenes are sourced from Landsat 8 and Landsat 9 Collection 2 Level-1 products, accessed through the USGS/EROS M2M API \cite{landsat8_9_c2_l1}. These products provide radiometrically calibrated surface observations, along with pixel level quality information (QA PIXEL band). To construct a representative and balanced training set, we employed a stratified sampling strategy, the results of which are illustrated in Figures~\ref{fig:landsat_sampling}, ~\ref{fig:climate_zones} and~\ref{fig:months}. Scenes were selected according to three main criteria: spatial diversity, temporal diversity, and cloud coverage. To ensure spatial diversity, land locations were randomly selected from a global coverage grid, ensuring balanced distribution across continents, climate zones, and elevation ranges. Temporal diversity was achieved by selecting scenes distributed approximately uniformly across all twelve months, capturing seasonal variations in vegetation, snow presence, solar angle, and atmospheric conditions. We focus on scenes with total reported cloud cover between 30 and 70$\%$, prioritizing partially cloudy conditions. For each selected cloudy scene, we retrieved a corresponding ``clear'' scene from the same location, with reported cloud cover below 5$\%$, captured as close in time as possible,  typically within 30 days, to serve as a cloud-free reference. The final sampling resulted in 617 cloudy and clear scene pairs, from which a total of 21,058 patches were extracted. In addition to the satellite observations, we augment each cloudy scene with a set of auxiliary geospatial layers that capture terrain, surface type, and atmospheric context. Specifically, we extract a DEM derived from the NOAA Global Relief Model 
\cite{etopo2v2} and a LULC layer informed by the Copernicus Global Land Cover 100m dataset
\cite{copernicus_lulc_2019}. Finally, we include five cloud related meteorological variables (surface pressure, dew point temperature, skin temperature, total precipitation and snow cover), obtained by ERA5 Land data \cite{era5land_muñoz2019}. Precipitation and snow cover values are log-transformed to reduce distribution skewness, facilitating normalization and stabilizing model training. All auxiliary bands are spatially aligned with the cloudy scene observation to ensure consistent pixel-level correspondence. These layers are included to provide additional spatial and environmental context for cloud interpretation.

All patches are resampled to a fixed resolution of 100 metres and cropped to 256×256 pixels, using average resampling, except for the weather, elevation and LULC layers which were resampled using nearest neighbor. Cloud masks are generated using the information in the QA PIXEL band. Pixels are classified as clear, thin cloud, or cloud based on the high confidence flags provided in QA PIXEL. To ensure a clean and processing-ready dataset, patches containing missing or invalid values in any spectral band or auxiliary layer were excluded during patch extraction.
Each one of the patches contains 39 bands. The 16 bands from the cloudy scene include Landsat bands B1–B11, spanning reflective and thermal infrared wavelengths, the QA PIXEL band, a cloud mask derived from QA PIXEL, the radiance values of Bands 10 and 11, and the mean radiance across Bands 10 and 11. The clear counterpart includes an identical set of 16 bands, allowing direct comparison with the corresponding cloudy scene. In addition, each patch includes DEM, LULC and the five meteorological bands described previously. 

\begin{figure*}[ht]
    \centering
    \includegraphics[width=\textwidth]{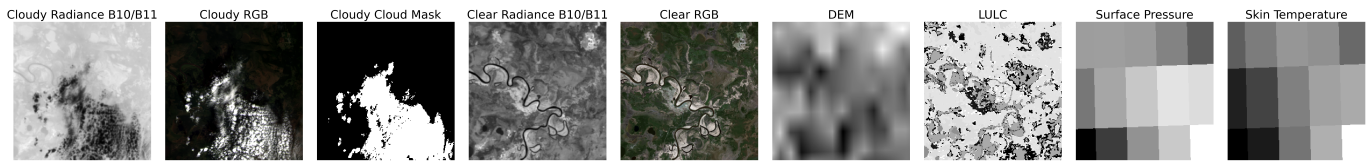}
    \caption{Example bands from a 256×256 Landsat 8 patch.}
    \label{fig:landsat_bands}
\end{figure*}

\Cref{fig:landsat_bands} shows example bands from a 256×256 patch, including cloudy and clear radiance (B10/B11), RGB composites, the corresponding cloud mask, and auxiliary layers such as DEM, LULC, surface pressure, and skin temperature.


\subsection{MA Landsat Dataset}


The \textit{MA Landsat} dataset is a complementary subset derived through patch extraction from the manually annotated cloud mask dataset described in \cite{scaramuzza_landsat8_col1_2021, scaramuzza_landsat8_col2_2021}. It primarily serves as an evaluation set, providing high-quality manual annotations for assessing model performance, and it can also be used for enhancing model performance. Compared to the \textit{Landsat Main} dataset, it is approximately an order of magnitude smaller.

The cloud annotations are sourced from the Landsat 8 Collection 1 and Collection 2 Cloud Truth Mask Validation Set, a human labeled dataset developed by USGS for the purpose of validating cloud detection algorithms \cite{scaramuzza_landsat8_col1_2021, scaramuzza_landsat8_col2_2021}. These annotations provide high precision cloud masks with pixel level labels for clear, thin cloud, and cloud classes. This subset is composed of 68 Landsat 8 scenes, from which we extract a total of 2,563 patches. In addition to the satellite observations, each patch is supplemented with auxiliary layers that mirror those used in the \textit{Landsat Main} set, including a DEM, a LULC map, and five meteorological variables. All auxiliary layers are spatially aligned with the annotated cloudy observation to ensure consistent and reliable context for model evaluation. Each patch in the \textit{MA Landsat} subset has the same size (256x256) resolution (100m) and includes the same 39 bands as those used in the Main Landsat Dataset, ensuring consistency in the input format for model training. 


\subsection{VIIRS Dataset}

The VIIRS Dataset is constructed to assess cloud detection model performance under nighttime imaging conditions, where visible reflectance is largely absent and thermal infrared observations become the primary source of information. This subset complements the daytime Landsat-based evaluation by specifically targeting challenging low-illumination scenarios. By doing so, it enables the evaluation of cloud segmentation models in environments where traditional optical signals are unavailable or unreliable, emphasizing robustness to lighting conditions.

\begin{figure}[htbp]
    \centering
    \includegraphics[width=1\linewidth]
    {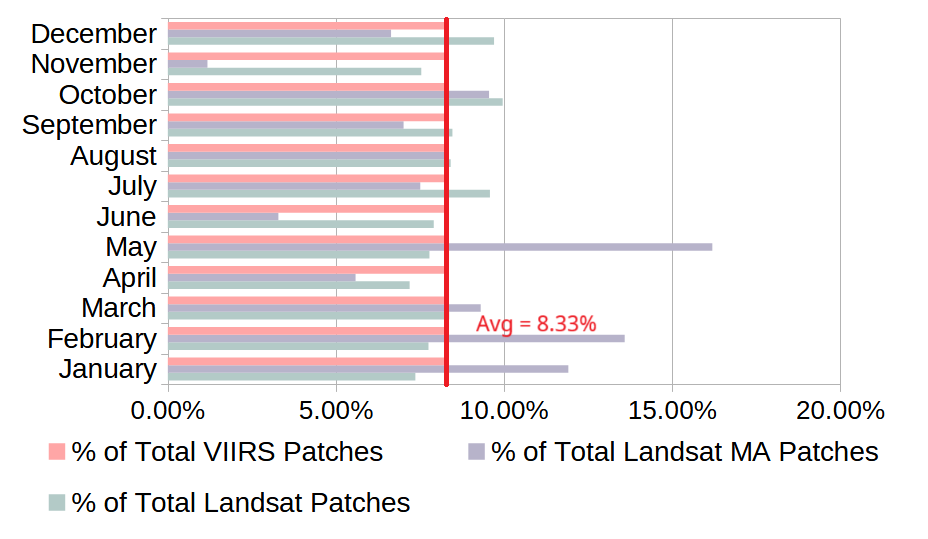}
    \caption{Temporal distribution of patches across months for VIIRS, Manually Annotated Landsat, and Landsat subsets.}
    \label{fig:months}
\end{figure}

To ensure broad temporal representation and geographic diversity, we select twelve orbital scenes from the Visible Infrared Imaging Radiometer Suite (VIIRS), one for each month of the year (Figure~\ref{fig:months}). These scenes are obtained from the Light Every Night (LEN) project, a public dataset compiled by the World Bank that provides daily VIIRS nighttime observations \cite{worldbank_len_2021}. Scenes are chosen to reflect a range of global environments, from urban to rural, tropical to polar, and across various atmospheric states. From these twelve segments, we extract a total of 5,004 image patches, providing a substantial and seasonally distributed test sample for model benchmarking.
Each patch contains five observational bands obtained directly from the LEN dataset. These include lunar illuminance (li), visible imagery (vis), thermal infrared brightness (tir), pixel sample position (samples), and a 16-bit bitmask (flag) that encodes quality and cloud information. In addition, each patch includes a derived binary cloud mask band, constructed by interpreting the flag bitmask and labeling pixels as either clear or cloud based on high confidence cloud indicators.

All patches are resampled to a uniform spatial resolution of 200 metres and cropped to a fixed size of 256 × 256 pixels. Following the same quality control procedure applied in the other two subsets, we exclude patches that contain missing values in any of the observational bands or that lack valid cloud mask information. Each patch in this subset is thus a valid nighttime observation, ready for use as input for model training and evaluation.

\section{Benchmark and Results}
\label{sec:bandr}

To demonstrate the utility of the proposed dataset, we design a comprehensive benchmark, assessing i) the potential of auxiliary features to enhance segmentation capabilities; and (ii) the dataset’s capacity to support the extraction of deep, cross-sensor / cross-domain features improving model transferability. Moreover, we offer qualitative results, showcasing how these features can enhance segmentation performance.

\subsection{Feature Combination Benchmarks}

 In Tables~\ref{tab:model_landsat_2_class} and~\ref{tab:model_landsat_3_class}, we present a series of experiments using different feature combinations from the Landsat Main dataset. From the first to the last row, we first show results from models trained with all relevant spectral Landsat bands known to contribute to cloud detection. Subsequently, we evaluate performance using only the thermal radiance band (TIR) and then progressively introduce combinations of the thermal band with various auxiliary inputs: the co-registered clear-sky reference (TIR/Clear), surface elevation (TIR/DEM), meteorological variables (TIR/Meteo), and finally all auxiliary data combined. The Landsat-based training dataset described in \Cref{sec:dd} is split into 70\% for training, 15\% for validation, and 15\% for testing. A strict group-split policy is applied, where groups correspond to scene IDs, to ensure that no patch from the same scene appears in different splits, thus mitigating spatial correlation. We employ four architectures to represent both generic and cloud-specialized semantic segmentation models. From the convolutional architecture family, we use the widely adopted \textit{U-Net} with a \textit{ResNet34} encoder \cite{ronneberger2015u} and the cloud-specific \textit{CDNetV2} \cite{cdnetv2}, a model tailored for cloud segmentation that incorporates channel and spatial attention mechanisms (AFFM) to enhance feature representations (see \Cref{sec:rw}). From the transformers family, we apply \textit{SegFormer} with a \textit{MiT-B2} encoder \cite{xie2021segformer} and \textit{SwinCloud}, a Swin-UNet adaptation for cloud detection that integrates a Global–Local Swin Transformer block (GLST) and a Feature Fusion Module (FFM) for multi-scale cloud detection in thermal infrared imagery (see \Cref{sec:rw}).
 
We present the Accuracy and mean Intersection over Union (mIoU) in \Cref{tab:model_landsat_2_class,tab:model_landsat_3_class} for the two-class (0: clear, 1: cloud) and three-class (0: clear, 1: thin cloud, 2: cloud) targets, respectively. Notably, thin clouds exhibit inherent ambiguity due to their semi-transparent nature, introducing aleatoric label uncertainty. This uncertainty hinders model training by generating inconsistent supervisory signals, thereby degrading performance \cite{kondylatos2025probabilistic}. As a result, mIoU scores for the three-class problem are consistently lower than those for the two-class setup across all related works. Our experiments indicate that certain feature combinations demonstrate improved performance (bold numbers in tables \ref{tab:model_landsat_2_class}, \ref{tab:model_landsat_3_class}) over the thermal-only baseline, particularly when using the attention-equipped SegFormer, CDNetV2 and SwinCloud. In contrast, the U-Net model does not appear to benefit from additional inputs. This tendency becomes more pronounced in the 3-class problem and in the combinations where the meteorology variables are included.  

\begin{table}[htbp]
\scriptsize
\centering
\caption{Model performance (2-Class Cloud Segmentation) across feature combinations}
\setlength{\tabcolsep}{2.5pt}
\label{tab:model_landsat_2_class}
\fontsize{9pt}{11pt}\selectfont
\begin{tabular}{l c c c c c c c c}
\toprule
\multirow{2}{*}{\textbf{Features}}& \multicolumn{2}{c}{\textbf{CDnetV2}} & \multicolumn{2}{c}{\textbf{SegFormer}} & \multicolumn{2}{c}{\textbf{U-net}} & \multicolumn{2}{c}{\textbf{SwinCloud}} \\
\cmidrule(lr){2-3} \cmidrule(lr){4-5} \cmidrule(lr){6-7} \cmidrule(lr){8-9} & \textbf{Acc} & \textbf{mIoU} & \textbf{Acc} & \textbf{mIoU} & \textbf{Acc} & \textbf{mIoU} & \textbf{Acc} & \textbf{mIoU} \\
\midrule
All Bands & 0.93 & 0.86 & 0.93 & 0.86 & 0.93 & 0.86 & 0.94 & 0.89 \\
TIR & 0.87 & 0.75 & 0.90 & 0.81 & 0.90 & 0.81 & 0.89 & 0.79 \\
TIR/Clear & 0.87 & \textbf{0.76} & 0.90 & 0.81 & 0.88 & 0.78 & 0.86 & 0.75\\
TIR/DEM & 0.86 & 0.74 & 0.90 & 0.81 & 0.90 & 0.80 & 0.86 & 0.74\\
TIR/Meteo & \textbf{0.89} & \textbf{0.79} & 0.90 & 0.81 & 0.90 & 0.80 & \textbf{0.90} & \textbf{0.80} \\
TIR/Aux. & \textbf{0.89} & \textbf{0.79} & 0.90 & 0.81 & 0.89 & 0.79 & \textbf{0.90} & \textbf{0.80} \\
\bottomrule
\end{tabular}
\end{table}

\begin{table}[htbp]
\scriptsize
\centering
\caption{Model performance (3-Class Cloud Segmentation) across feature Combinations}
\label{tab:model_landsat_3_class}
\fontsize{9pt}{11pt}\selectfont
\setlength{\tabcolsep}{2.5pt}
\begin{tabular}{l c c c c c c c c}
\toprule
\multirow{2}{*}{\textbf{Features}} & \multicolumn{2}{c}{\textbf{CDnetV2}} & \multicolumn{2}{c}{\textbf{SegFormer}} &\multicolumn{2}{c}{\textbf{U-net}} & \multicolumn{2}{c}{\textbf{SwinCloud}} \\
\cmidrule(lr){2-3} \cmidrule(lr){4-5} \cmidrule(lr){6-7} \cmidrule(lr){8-9}
& \textbf{Acc} & \textbf{mIoU} & \textbf{Acc} & \textbf{mIoU} & \textbf{Acc} & \textbf{mIoU} & \textbf{Acc} & \textbf{mIoU} \\
\midrule
All Bands & 0.91 & 0.71 & 0.89 & 0.66 & 0.89 & 0.65 & 0.90 & 0.70\\
TIR & 0.83 & 0.53 & 0.86 & 0.56 & 0.86 & 0.58 & 0.80 & 0.52 \\
TIR/Clear & \textbf{0.84} & 0.53 & 0.86 & \textbf{0.59} & 0.84 & 0.55 & 0.80 & 0.50\\
TIR/DEM & 0.83 & 0.51 & 0.84 & \textbf{0.57} & 0.85 & 0.56 & 0.79 & 0.49 \\
TIR/Meteo & \textbf{0.87} & \textbf{0.54} & \textbf{0.88} & \textbf{0.60} & 0.86 & 0.55 & \textbf{0.83} & \textbf{0.55} \\
TIR/Aux. & \textbf{0.86} & \textbf{0.54} & \textbf{0.87} & \textbf{0.57 }& 0.86 & 0.57 & \textbf{0.86} & \textbf{0.55}\\
\bottomrule
\end{tabular}
\end{table}

\subsection{Visual examples}

The following examples illustrate how different auxiliary inputs resolve specific typical errors of TIR-only cloud detection. By contrasting the model’s outputs from cloudy TIR alone against those produced when clear TIR, DEM, and meteorological auxiliary data are used, we highlight how each auxiliary source provides complementary information that helps the model avoid common confusions.

\begin{figure*}[ht]
    \includegraphics[width=\textwidth]{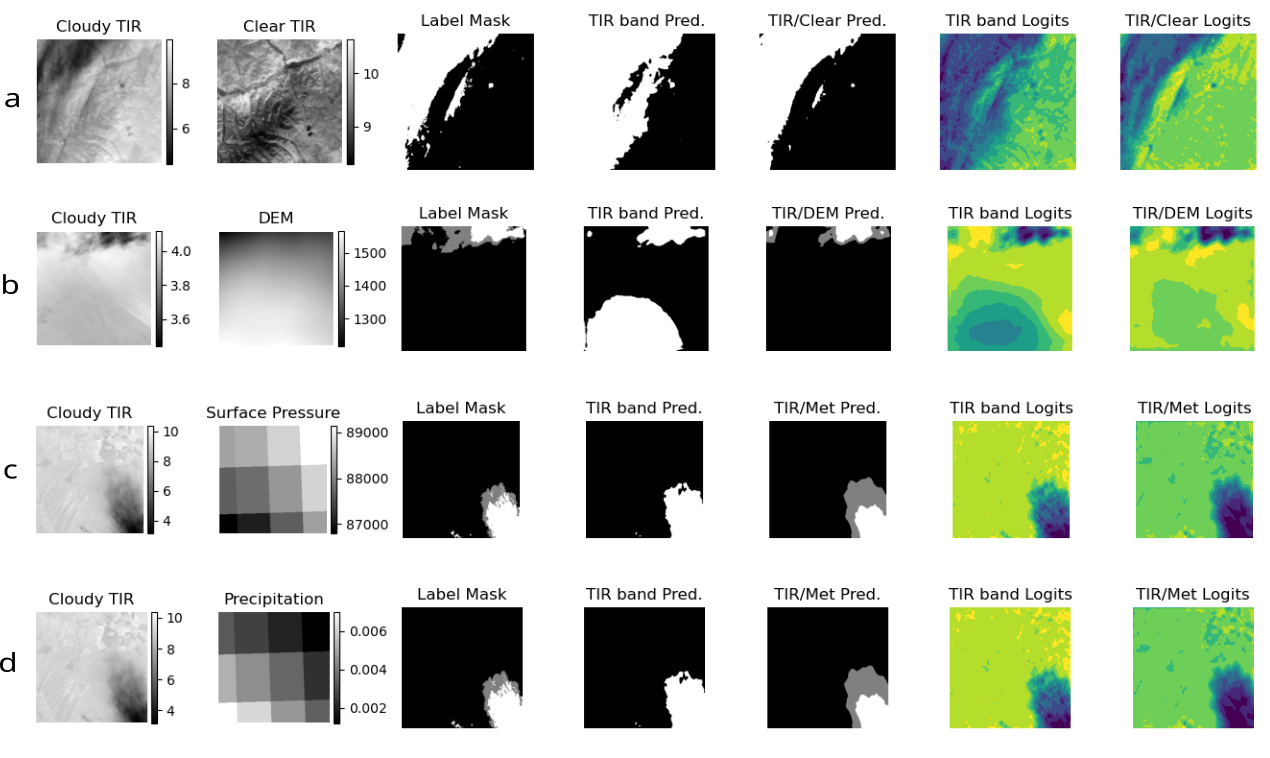}
    \centering
    \caption{Examples of inference where  auxiliary input contributed to the improvement of accuracy of the model. (a) The model trained only on Cloudy TIR predicts a false positive because the ground in that area exhibited lower than usual thermal values and real clouds were present nearby, causing the region to resemble a cloud. Clear TIR helped resolve this uncertainty. (b) The inclusion of the DEM helps resolve the common infrared based confusion between clouds and high altitude cold terrain (ice/snow), evident in the TIR-only prediction that mistakenly labels part of a mountain as cloud. (c) and (d) The combination of low precipitation and high surface pressure (conditions that suppress deep moist convection and the formation of thick clouds) helps the model more effectively distinguish thin cloud layers from denser cloud cover.}
    \label{fig:inference}
\end{figure*}

In \Cref{fig:inference}a, the clear image reveals a part of a mountainous area where the model trained only on Cloudy TIR incorrectly detects a cloud. The additional input provided to the model trained with both Cloudy and Clear TIR contributes to recognizing the clear ground, whereas the Cloudy TIR-only model fails to distinguish between the lower thermal radiance and the presence of cloud structures.

In \Cref{fig:inference}b, the inclusion of the DEM helped reduce confusion between clouds and elevated terrain. As seen in the related image, the prediction made by using only the TIR band incorrectly labels part of a high mountain region as cloud. This is a common challenge in infrared based cloud detection, where cold surface temperatures at high altitudes can mimic the radiative signature of cloud tops. Once DEM is incorporated, the model correctly identifies this area as terrain rather than cloud, significantly improving prediction accuracy.

The combination of low precipitation and high surface pressure, as seen in Figures \ref{fig:inference}c and \ref{fig:inference}d, can play a crucial role in improving the model’s ability to differentiate thin clouds from thicker, high confidence cloud cover. High surface pressure is typically associated with stable atmospheric conditions, which inhibits the formation of thick clouds. At the same time, low precipitation levels further support the absence of deep moist convection, favoring the presence of thinner cloud layers.

\subsection{Cross-Domain transferability}
To demonstrate the utility of the Landsat Main dataset as a source domain, we perform transfer learning experiments by adapting the learned representations to the MA Landsat and VIIRS datasets. In the case of the \textit{MA Landsat} dataset, the target masks differ slightly because they are manually annotated, resulting in a minor domain shift. In contrast, the \textit{VIIRS} dataset exhibits a much stronger domain shift, as its TIR band has an original spatial resolution of 750 m (compared to 100 m for Landsat) and originates from a different sensor with slightly shifted spectral wavelengths. \Cref{tab:fine-tune} showcases the results of the experiments. 

Following the order of rows in \Cref{tab:fine-tune}, models were first trained using the \textit{U-Net} and \textit{SegFormer} architectures directly on \textit{VIIRS} and \textit{MA Landsat} datasets to establish baselines. Next, we used the models pretrained on \textit{Landsat Main} to perform a series of transfer learning experiments. We ran inference with the pretrained models and then fine-tuned them on \textit{VIIRS} and \textit{MA Landsat} dataset by fully or partially freezing the encoder. In the partial-freeze case, the first three layers of the \textit{U-Net} and the overlapping patch embedding stages of the \textit{SegFormer} were frozen. All models were evaluated on the test splits of the \textit{VIIRS} and \textit{MA Landsat} datasets.

The results in Table~\ref{tab:fine-tune} show that, across all transfer learning experiments, even with no fine-tuning, models pretrained on \textit{Landsat Main} consistently outperform those trained directly on \textit{MA Landsat}. In the case of \textit{VIIRS}, only the SegFormer fine-tuned models with partially frozen encoders pretrained on \textit{Landsat Main} surpass the performance of models trained directly on \textit{VIIRS}.

These findings demonstrate that the \textit{Landsat Main} dataset serves as a strong source domain for learning transferable representations to \textit{MA Landsat}, which has more accurate labels.  An important observation is that models pretrained on Landsat Main achieve better performance even without fine-tuning.
However, transferring to a different sensor requires substantial fine-tuning, which nevertheless tends to yield improved performance compared to training solely on the target dataset.


\begin{table}[htbp]
\centering
\caption{Training and fine-tuning results on \textit{VIIRS} and \textit{MA Landsat} datasets using Landsat-Main pretraining. Thermal band is used for training.}
\label{tab:fine-tune}
\setlength{\tabcolsep}{2pt}
\fontsize{9pt}{11pt}\selectfont
\begin{tabular}{p{1.6cm} p{3cm} c c c c}
\toprule
\multirow{2}{*}{\textbf{Dataset}} & {\textbf{Training scheme}} & \multicolumn{2}{c}{\textbf{U-Net}} & \multicolumn{2}{c}{\textbf{SegFormer}} \\
\cmidrule(lr){3-4} \cmidrule(lr){5-6}
& FT: Fine Tune & \textbf{Acc} & \textbf{mIoU} & \textbf{Acc} & \textbf{mIoU} \\
\midrule
\multirow{4}{*}{VIIRS}
& Train & 0.79 & 0.65 & 0.80 & 0.66 \\
& Pretrained Inference & 0.43 & 0.22 & 0.45 & 0.24 \\
& FT - Frozen encoder & 0.77 & 0.63 & 0.78 & 0.63 \\
& FT - Partially Frozen & 0.78 & 0.64 & 0.80 & \textbf{0.67} \\
\midrule
\multirow{4}{*}{MA Landsat}
& Train & 0.87 & 0.77 & 0.84 & 0.72 \\
& Pretrained Inference & \textbf{0.89} & \textbf{0.80} & \textbf{0.86} & \textbf{0.75} \\
& FT - Frozen encoder & \textbf{0.89} & \textbf{0.79} & \textbf{0.86} & \textbf{0.75} \\
& FT - Partially Frozen & \textbf{0.88} & \textbf{0.78} & \textbf{0.86} & \textbf{0.76} \\
\bottomrule
\end{tabular}
\end{table}
\section{Discussion}
\label{sec:disc}

TIRAuxCloud addresses the lack of datasets supporting the development of models for continuous, day-and-night cloud detection, which is a critical task for a variety of important remote sensing services. It provides a comprehensive yet computationally practical dataset suite. The dataset combines large-scale automatically labeled subsets with manually annotated data. It also includes auxiliary features that are available during both day and night. These features are related to cloud formation and atmospheric conditions. Together, they complement thermal imaging and help improve cloud detection models. The benchmark results confirm the rich informativeness of the dataset compared to other cloud segmentation studies (e.g. CDNetV2 \cite{cdnetv2}, CloudU-Net \cite{cloudu-net}, Cloud-Net+ \cite{95-cloud}, SwinCloud \cite{swincloud}).

Our experiments with auxiliary feature combinations showed that attention-equipped models tend to benefit from these additional inputs. However, this improvement remains limited, though it appears consistent across certain feature combinations and architectures. While it is well established that the information contained in auxiliary variables can help reduce model uncertainty and ambiguity, integrating them effectively into a learning framework remains nontrivial. A possible reason for the model’s difficulty in exploiting these sources is \textit{unimodal dominance} \cite{multimodalbias}, the tendency of one modality (in this case, the thermal band) to dominate learning. Since the thermal band is closely correlated with the target variable (the cloud mask), the model tends to prioritize this simpler, high-signal pathway while neglecting more complex multimodal relationships.
Achieving better performance may require problem-specific architectural solutions. For instance, difference based feature extractors could be employed to explicitly model spectral spatial deviations between cloudy and co-registered clear-sky counterparts. Similarly, coarse resolution meteorological variables could be integrated through a parallel encoder branch operating at lower spatial frequencies.

Experiments on the \textit{MA Landsat} dataset showed that models pretrained on \textit{Landsat Main} outperformed those trained directly on the MA Landsat. Specifically, the pretrained U-Net and SegFormer models outperform those trained on \textit{MA Landsat} by 0.03 in mIoU and by 0.02 in accuracy. This suggests that for cloud segmentation, large-scale automatically labeled data with moderate label noise (e.g., operational Landsat cloud mask products) may be more valuable than smaller manually annotated datasets, indicating that dataset scale can, in some cases, outweigh annotation fidelity.

In contrast, transfer learning to the \textit{VIIRS} dataset requires substantial fine-tuning due to the larger domain gap, yet still achieves comparable or slightly better results than models trained exclusively on \textit{VIIRS}. 
More work is needed to strengthen cross-sensor generalization, for example by augmenting the dataset with additional and more diverse sensor sources, leveraging self-supervised and foundation model pretraining, and exploring domain adaptation strategies. 

\textbf{Future work directions:}
\begin{itemize}
  \item Extend dataset coverage across additional sensors, modalities, and geographic regions to enhance generalization.
  \item Develop models capable of mitigating unimodal dominance to better exploit auxiliary information and improve overall performance.
  \item Develop self-supervised pretraining strategies for more robust feature learning.
  \item Explore lightweight architectures optimized for in-orbit deployment.
  \item Evaluate foundation models on the dataset to assess transferability and cross-domain adaptation.
  \item Apply explainability techniques to better understand the contribution of auxiliary features.
\end{itemize}

\section{Conclusion}
\label{sec:concl}
We presented TIRAuxCloud, a novel multi-modal dataset for cloud segmentation. The main contributions include:
i) the creation of a large-scale dataset enriched with comprehensive auxiliary data, designed as a testbed for advancing methods in day- and night-time cloud detection;
ii) benchmark evaluations using both general-purpose and cloud-specialized models, which establish strong performance baselines and demonstrate the dataset’s potential in improving cloud detection methods; and
iii) insights into how auxiliary features can contribute to enhanced model accuracy.
Further research and experimentation are needed to fully exploit the dataset’s potential and ultimately improve the accuracy of critical remote sensing applications operating continuously, day and night.

\bibliographystyle{IEEEtran}
\bibliography{main.bib}

\end{document}